\documentclass{article}
\usepackage{ijcai19}

\usepackage{times}
\usepackage{soul}
\usepackage{url}
\usepackage[hidelinks]{hyperref}
\usepackage[utf8]{inputenc}
\usepackage[small]{caption}
\usepackage{graphicx}
\usepackage{subfigure}
\usepackage{amsmath}
\usepackage{booktabs}
\usepackage{algorithm}
\usepackage{algorithmic}
\usepackage{multirow}
\usepackage{enumitem}
\title{Adaptive Portfolio by Solving Multi-armed Bandit via Thompson Sampling}

\author{
Mengying Zhu$^1$\and
Xiaolin Zheng$^1$\footnote{Contact Author}\and
Yan Wang$^2$\And
Yuyuan Li$^1$ and
Qianqiao Liang$^1$\\
\affiliations
$^1$Department of Computer Science, Zhejiang University, Hangzhou, China\\
$^2$Macqaurie University, Department of Computing, Sydney, NSW, Australia\\
\emails
\{mengyingzhu,xlzheng,11821022,liangqq\}@zju.edu.cn,
yan.wang@mq.edu.au
}

\begin{document}
\nocite{*}
\maketitle

\begin{abstract}
As the cornerstone of modern portfolio theory, Markowitz's mean-variance optimization is considered a major model adopted in portfolio management. However, due to the difficulty of estimating its parameters, it cannot be applied to all periods. In some cases, naive strategies such as Equally-weighted and Value-weighted portfolios can even get better performance. Under these circumstances, we can use multiple classic strategies as multiple strategic arms in multi-armed bandit to naturally establish a connection with the portfolio selection problem. This can also help to maximize the rewards in the bandit algorithm by the trade-off between exploration and exploitation. In this paper, we present a portfolio bandit strategy through Thompson sampling which aims to make online portfolio choices by effectively exploiting the performances among multiple arms. Also, by constructing multiple strategic arms, we can obtain the optimal investment portfolio to adapt different investment periods. Moreover, we devise a novel reward function based on users' different investment risk preferences, which can be adaptive to various investment styles. Our experimental results demonstrate that our proposed portfolio strategy has marked superiority across representative real-world market datasets in terms of extensive evaluation criteria.
\end{abstract}

\section{Introduction}
The portfolio selection problem is a fundamental issue in the financial sector for many asset investments, including funds, stocks, bonds, and options. According to Gary Brinson, the father of global asset allocation,  ``Asset allocation is the main factor that affects all overall returns.'' In the long run, more than 90\% of a portfolio's performance is attributable to its asset allocation  \cite{brinson1995determinants}. Thus, asset allocation of a portfolio is the key determinant of performance, risk, and volatility over time. 

Modern portfolio theory and analysis tend to build upon the seminal work of Markowitz  \cite{markowitz1952portfolio}. Up to now, the mean-variance paradigm has remained the mainstream choice for academia and industry. However, the main problem of using a single strategy is that it cannot be adapted to the changing environment. For instance, in an event of the stock market crash, the sold-all strategy with cash will have a good performance. Meanwhile,  during the bull market, the buy-and-hold strategy is likely to perform even better. In terms of this issue, a simple approach in the investment field is to periodically review the effectiveness of the current strategy and appropriately adjust the strategy for the next phase. This is a typical problem of exploration and exploitation. Therefore, we use reinforcement learning to solve the problem of how to determine optimal portfolio strategy to adapt for different investment periods.

Meanwhile, each investor's pursuit of risk and benefit is different, which is called the user's investment risk preference. Although a good strategy can ultimately help in achieving a good return, not every investor is willing to take on some risks in the process. For example, users with a low risk tolerance are unlikely to consider short-term losses, whereas users with high risk tolerance tend to pursue high returns and usually do not care about retracement. For this reason, we take the users' investment risk preferences into account and propose a novel reward function, which can be adaptive to various investment styles such as high-risk-high-return and low-risk-low-return investment style.

In this paper, we first turn the portfolio problem into a multi-armed bandit problem and construct a series of strategic arms basing on the classic strategies. Subsequently, we apply Thompson sampling method to select strategic arm and further update the Beta distribution of strategic arms based on the user's investment risk preference. The contributions of the present work are summarized as follows:
\begin{itemize}
	\item To adapt to different market conditions in different periods, we utilize the multi-armed bandit problem to adaptively select the most suitable strategy to form an online portfolio strategy.
	\item We devise a novel reward function based on the users' investment risk preferences to ascertain that the method fits a variety of users' needs. This helps to achieve different return-to-risk ratio. 
	\item Experimental results indicate that the proposed portfolio strategy has marked superiority across representative real-world market datasets in terms of a series of standard financial evaluation indicators, which include Sharpe ratios, cumulative wealth, volatility, and maximum drawdowns. 
\end{itemize}

\section{Related Work}
In this section, we briefly discuss two topics, that is, multi-armed bandit and portfolio selection problem. 

\subsection{Multi-armed Bandit and Thompson Sampling}
This section contains theories, solutions for the multi-armed bandit problem and Thompson sampling.

 There are many exploration vs exploitation dilemmas in many aspects of our life. At the same time,  investment strategies attempt to balance existing portfolios and new portfolios to achieve higher returns. In this case, if we can speculate the future trend of all assets in the market, we can find the best investment strategy by just simulating brute-force instead of using several other smart approaches. This dilemma originates from the incomplete information: we need to gather enough information to make best overall decisions while keeping the risk under control. With exploitation, we can take advantage of the best known option. With exploration, we can take some risk to collect information about unknown options. Therefore, the best long-term strategy may involve short-term sacrifices.

The multi-armed bandit problem is a classic problem that exhibits the exploration vs exploitation dilemma. It is like facing multiple slot machines in a casino and each is configured with an unknown probability of how likely you can get a reward at one play. The aim is to maximize the cumulative reward. If we know the optimal action with the best reward, then the goal is same as to minimize the potential regret or loss by not picking the optimal action.

The possible methods that can be used to solve this problem are roughly divided into three distinct  categories, $\epsilon$-greedy algorithm, upper confidence bounds (UCB) algorithm  \cite{auer2002finite} and Thompson sampling \cite{thompson1933likelihood}.

Thompson sampling has a simple idea. However, it works great for solving the multi-armed bandit problem  \cite{chapelle2011empirical,russo2014learning}. At each time step, select action $a$ according to the Beta probability that $a$ is optimal. After observing the true reward, update the Beta distribution accordingly. This essentially involves doing Bayesian inference to compute the posterior with the known prior and the likelihood of getting the sampled data. 

With the rise of reinforcement learning, numerous works study how to apply multi-armed bandit to various fields, such as recommender \cite{li2010contextual,wu2016contextual} and e-commence \cite{broden2017bandit,broden2018ensemble}. Besides, some scholars have tried to incorporate reinforcement learning into the field of portfolio optimization  \cite{Liang2018Adversarial,jiang2017deep,sani2012risk}. Still, other studies have used assets directly as arms in multi-armed bandit. For instance, Shen \cite{shen2015Portfolio} proposed to use the UCB algorithm to achieve online portfolio selection by constructing an orthogonal portfolio. Meanwhile, other studies have only used Thompson sampling to generate portfolio. For example, Shen \cite{shen2016portfolio} presented an online portfolio algorithm that leverages Thompson sampling to mix two different strategies. Inspired by these studies, we combine the multi-armed bandit and Thompson sampling, use the classic strategies as strategy arms to achieve an adaptive portfolio.

\subsection{Portfolio Strategy}
This section presents the current state of research on portfolios, including mean-variance models, forecast trends, and the Universal portfolio. Existing studies have specially based upon classical financial theory and have combined with machine learning to achieve better performance.

In 1952, Markowitz put forward the mean-variance model, which was the first of its kind in modern portfolios \cite{markowitz1952portfolio}. This model constrains the relevant conditions of portfolio issues to pursue a balance of risk and return. In particular, some scholars attempted to improve the effect of the mean-variance model by adding regularity \cite{brodie2009sparse,shen2014doubly}. Other studies have improved the performance of the mean-variance model by changing the sampling method. For instance, Shen \cite{shen2017portfolio} proposed a new portfolio strategy through resampling subsets of the original large universe of assets.

In addition, some scholars pursued the maximum return-to-risk ratio of the portfolio through trend forecasting, such as by predicting stock price movements in the stock market. For example, Palmowski et al. \cite{palmowski2018optimal} studied a portfolio selection problem in a continuous-time It\^{o}-Markov additive market in which the prices of financial assets were described by Markov additive processes. Meanwhile, Paolinelli \cite{paolinelli2019model} proposed a model for stocks dynamics based on a non-Gaussian path integral, which connected between time horizons and trading strategies.

The third type of research is based on the Universal portfolio theory. This is a portfolio selection algorithm from the field of machine learning and information theory. The algorithm learns adaptively from historical data and maximizes the log-optimal growth rate in the long run. Huang et al. \cite{huang2015semi} designed semi-universal portfolio strategy under transaction fee, which tries to avoid rebalancing when the transaction fee outweighs the benefit of trading.

All of the above methods are all based on a single financial theory to construct an online investment portfolio. However, the method of this paper adaptively adopts different investment strategies in multiple cycles to achieve the highest long-term return-to-risk ratio.

\section{Methodology}
In this section, we first introduce the notations and finance terms used in this paper. We will also discuss several strategic arms based on classic portfolios, formulate portfolio blending a multi-armed bandit problem, and how to solve this problem using Thompson sampling. Lastly, we summarize the proposed algorithm.

\subsection{Notations and Problem Definition}
To start with, we give the problem an abstract definition. We consider a self-financing, limited time and  limited asset financial environment. The trading periods consist of $t_{k} = k \Delta t, k =0, ... , m$, where $\Delta t$ represents one day, week or month, depending on the cycle of rebalancing and $m$ is the total cycles of participation in the transaction. We also represent the return vector of $n$ assets at time $t_{k-1}$ to $t_{k}$ time as $\mathbf{R_{k}}$. The formula of the return $R_{k,i}$ of the i-th asset is $R_{k,i} = P_{k,i}/P_{k-1,i}$, where $P_{k-1,i}$ and $P_{k,i}$ represent the price of the i-th asset at times $t_{k-1}$ and $t_{k}$. The transaction fee is also an important factor in the final benefit. For the sake of simplifying the model, however, it is not considered in this model. Still, we think about how to reduce trading behavior.

$\mathbf{W_{k}}$ as the portfolio weight vector at time $t_{k}$ denotes the investment decision at time $t_{k}$, where $W_{k,i}$ represents the allocation weight of the i-th asset in the entire portfolio. We assume that the sum of the combined weights is $1$ (except for pure cash position), i.e., $\mathbf{W_{k}^{T}}\mathbf{1} = 1$, where $\mathbf{1}$ is a column vector with ones as its entities. Also, we correspond to the following two cases of $\mathbf{W_{k}}$ and the actual trading strategy: $w_{k,i} > 0$ indicates that we need to take a long position of the i-th asset at market price; while $w_{k,i} < 0$ shows that we need to take a short sale position of i-th asset. The actual operation requires a deposit, and also needs to pay dividends for short-selling assets, etc. However, for the sake of simplification, we will not consider this situation for the time being, and only consider the gains or losses caused by stock price changes.

\subsection{Strategic Arms Based on Classic Portfolios}

In our research, we do not directly use assets as arms in multi-armed bandit. Instead, we use classic portfolio strategies in finance as strategic arms to reduce the number of arms, and also to reduce transaction volume as well as increase stability. We use the following strategies:

\textbf{Buy and Hold (BH):}
This is an intuitive idea which involves doing nothing and continuing to hold the existing portfolio in this time window.
\begin{equation}
\mathbf{W_{k}^{BH}}=\mathbf{W_{k-1}} \label{con:BH}.
\end{equation}

\textbf{Sold All (SA):}
Involves selling all the assets so that the combination is an empty position or a pure cash position.
\begin{equation}
\mathbf{W_{k}^{SA}}=\mathbf{0} \label{con:SA}.
\end{equation}

\textbf{Equally-weighted portfolio (EW):}
Regardless of the asset, all assets are directly placed into equal weight positions during each rebalancing period.
\begin{equation}
\mathbf{W_{k}^{EW}}=\frac{1}{n}\mathbf{1} \label{con:EW}.
\end{equation}

\textbf{Value-weighted portfolio (VW):}
As a passive investment strategy, positions in each rebalancing period are allocated as per the current capital of each asset.
\begin{equation}
\mathbf{W_{k}^{VW}}=\frac{\mathbf{W_{k-1}}\circ \mathbf{R_{k-1}}}{\mathbf{W_{k-1}^{T}} \mathbf{R_{k-1}}} \label{con:VW}.
\end{equation}

\textbf{Mean-variance portfolio (MV):}
Mean-variance model is a strategy constructed in line with the Markowitz's theory. It captures the aforementioned risk-return trade-off.
\begin{equation}
\mathbf{W_{k}^{MV}}=\arg \min _{\mathbf{W_{k}}\mathbf{1}=1}\mathbf{W_{k}^{T}}\mathbf{\Sigma _{k}}\mathbf{W_{k}}-\mathbf{R_{k}^{T}}\mathbf{W_{k}} \label{con:MV},
\end{equation}
where $\mathbf{R_{k}^{T}}\mathbf{W_{k}}$ is the expected return and $\mathbf{W_{k}^{T}}\mathbf{\Sigma _{k}}\mathbf{W_{k}}$ is the variance of portfolio returns.

\subsection{Portfolio Bandit via Thompson Sampling (PBTS)}
Each strategy has its own suitable period and scene, thus they also have a certain probability to get the most profit. Basing on this idea, this paper regards the portfolio selection problem as a multi-armed bandit problem, and classic portfolio strategies as the strategic arms in order to achieve higher long-term returns. The specific definition is as follows:

The multi-armed bandit of the portfolio strategy is $< a;R_{a} >$. $a$ is a collection of strategic arms (classic portfolio strategies),
	\begin{align}
	a_{k}=[a_{k,1},...,a_{k,l}],\nonumber \\
	a_{k,1}=w_{k}^{BH},a_{k,2}=w_{k}^{SA},a_{k,3}=w_{k}^{EW},\nonumber \\
	a_{k,4}=w_{k}^{VW},a_{k,5}=w_{k}^{MV},
	\end{align}
where $l$ represents the total number of strategic arms. There are $l$ arms at each time $k$, and which arm is selected according to which strategy is used to adjust the weight of the portfolio.

Assume $R_{a_{j,r}} = P(r|a_{j})$ is the probability distribution function of the return, at each time $k$, $\theta_{k,j} \sim R_{a_{j,r}}$. And the probability of each strategic arm is a Beta distribution
$\theta_{j} \sim Beta(\alpha _{j}, \beta _{j})$.

At time k, each arm randomly samples a value $\theta _{k, j}$ from its respective Beta distribution, then the arm $j_{k}$ of this selection is:
	\begin{equation}
	j_{k}=\arg \max_{j} \theta_{k, j}
	\label{con:Select}.
	\end{equation}
	
In order to judge whether this choice is successful, we comprehensively consider the users' investment risk preferences and use the Sharpe ratio as a measure. Therefore, we give a $(0,1)$ criterion based on the top-k strategy. The judgment formula is:
	
\begin{equation}
\begin{split}
		&\left\{\begin{matrix}
		& \sum_{j=1}^{l} (\mathbf{1}_A) \geq c &success \\ 
		&\sum_{j=1}^{l} (\mathbf{1}_A) <  c & failure \\ 
		\end{matrix}\right.  \\
		& A=\left \{j \mid  \right.SR(a_{k, j_k})-SR(a_{k,j}) \geq 0\left.  \right \},
\end{split}
\label{con:RF}
\end{equation}		
	where $\mathbf{1}_A$ is an indicator function and $SR(a_{k,j})$ represents the Sharp ratio of user's historical selection of arm $j$ at time $t_k$. Usually, the international average generally takes a 36-month net growth rate to calculate the Sharpe ratio. 
	
	The choice of $c$ can be selected based on users' investment risk preferences. If the user prefers to pursue high-risk and high-return, the smaller the $c$ can be; the larger the $c$ can be, if the user tends to pursue a relatively stable investment. 

Then update the Beta distribution of arm $j_{k}$, expressed as:
\begin{equation}
	\left\{\begin{matrix}
	 success & \theta_{j_{k}} \sim Beta(\alpha_{j_{k}}+1,\beta_{j_{k}})\\ 
	 failure & \theta_{j_{k}} \sim Beta(\alpha_{j_{k}},\beta_{j_{k}}+1)
	\end{matrix}\right.
	\label{con:SF},
\end{equation}
where $success/failure$ is determined by Equation (\ref{con:RF}).

Additionally, for each arm's Beta distribution, we first use $Beta (1, 1)$ as the initial prior of each arm and update the a priori results using sliding window $\tau $ of historical data. Since there is no investment strategy performance, $Beta (1, 1)$, the even distribution of standards, is a reasonable initialization for investors. At each rebalancing time, the investor builds the Bernoulli test described above, observes subsequent successes or failures, and updates the posterior distribution accordingly.

Algorithm 1 summarizes the process of building a multi-armed bandit problem and solving problem via Thompson sampling.

\begin{algorithm}[tb]
\caption{Portfolio Bandit via Thompson Sampling}
\label{alg:algorithm}
\textbf{Input}: Total cycles of participation in the transaction ($m$), number of asserts ($n$), daily return ($\mathbf{R}$), sliding window ($\tau $), the top ($c$) \\
\textbf{Output}: Portfolio weight ($\mathbf{w})$
\begin{algorithmic}[1] 
\STATE Initialize the Beta distribution $\theta_j \sim Beta (\alpha_j, \beta_j)$ of each strategic arm by $\alpha_1=...=\alpha_l=\beta_1=\beta_l=1$.
\FOR{ $k$ = $1$ to $m$} 
\STATE Calculate the weight ratio of each basic portfolio strategy according to Eqs. (\ref{con:BH}) - (\ref{con:MV}).
\STATE Sampling each arm's $\theta_{j, k}$ from the $Beta(\alpha_j, \beta_j)$ distribution .
\STATE Select arm $j_k$ according to Equation (\ref{con:Select}).
\IF {$k>\tau$}
\STATE Assign the portfolio weight $w_{k}=a_{k, j_{k}}$ at $t_k$.
\ENDIF
\STATE Update $\alpha_j$ and $\beta_j$ according to Eqs. (\ref{con:RF})-(\ref{con:SF}).
\IF {Success}
\STATE $\alpha_j=\alpha_j+1$.
\ELSE
\STATE $\beta_j=\beta_j+1$.
\ENDIF
\ENDFOR
\end{algorithmic}
\end{algorithm}

\section{Experiments}

\subsection{Data}

\begin{table*}
\centering
\begin{tabular}{lllllll}
\hline
Dataset  & Frequency & Time Period & m & n & Description \\
\hline
FF25       & Monthly  & 06/01/1963 - 11/31/2018  & 545  & 25   & 25 portfolios of firms sorted by size and book-to-market \\
FF49       & Monthly  & 07/01/1969 - 11/31/2018 & 472  & 49   & 49 industry portfolios representing the U.S. stock market \\
FF100    & Monthly  & 07/01/1963 - 11/31/2018  & 544  & 100  & 100 portfolios of firms sorted by size and book-to-market \\
ETFs   & Daily  & 12/08/2011 - 11/10/2017  & 1,138  & 608  & Exchange-traded funds in U.S. stock market \\
SP500   & Daily  & 02/11/2013 - 02/07/2018 & 1,355  & 476  & 500 firms
listed in the S\&P 500 Index \\
\hline
\end{tabular}
\caption{Summary of the datasets}
\label{tab:datasets}
\end{table*}

In our experiment, we consider two types of datasets. The first one is the FF dataset, which was built by Fama and French based on the US stock market and continues to be updated to date \cite{fama1992cross}. Overall, they have an extensive coverage of assets classes and span a long period. In our experiments, the FF25, FF49, FF100 datasets include monthly returns of 25, 49, and 100 assets more than half a century. Among them, FF25 and FF100 are formed on size and book-to-market, while FF49 is an industry portfolio. The second one is a more frequent stock market data, which includes constituents of the SP500 and ETFs in the US stock market. We exclude assets with missing data for the past five years. Thus, we remain with 476 stocks from 500 constituent stocks as well as 608 ETFs retained by 1,340 ETFs.

Table \ref{tab:datasets} is a summary of the datasets, representing different investment perspectives in the market. The FF datasets emphasize long-term gains, spanning more than half a century. They include the different periods of the US stock market as well as multiple financial crises that can reflect the long-term gains of the strategy. Meanwhile, the SP500 and ETF datasets reflect at high trading frequencies. Regardless of the extreme market, the medium-term performance of the strategy is highlighted. In particular, we choose the timing of our datasets to avoid the latest financial crisis after 2007.

\begin{figure*}[t]
\centering
\subfigure[FF25]{
\begin{minipage}[t]{0.33\linewidth}
\centering
\includegraphics[width=2.3in]{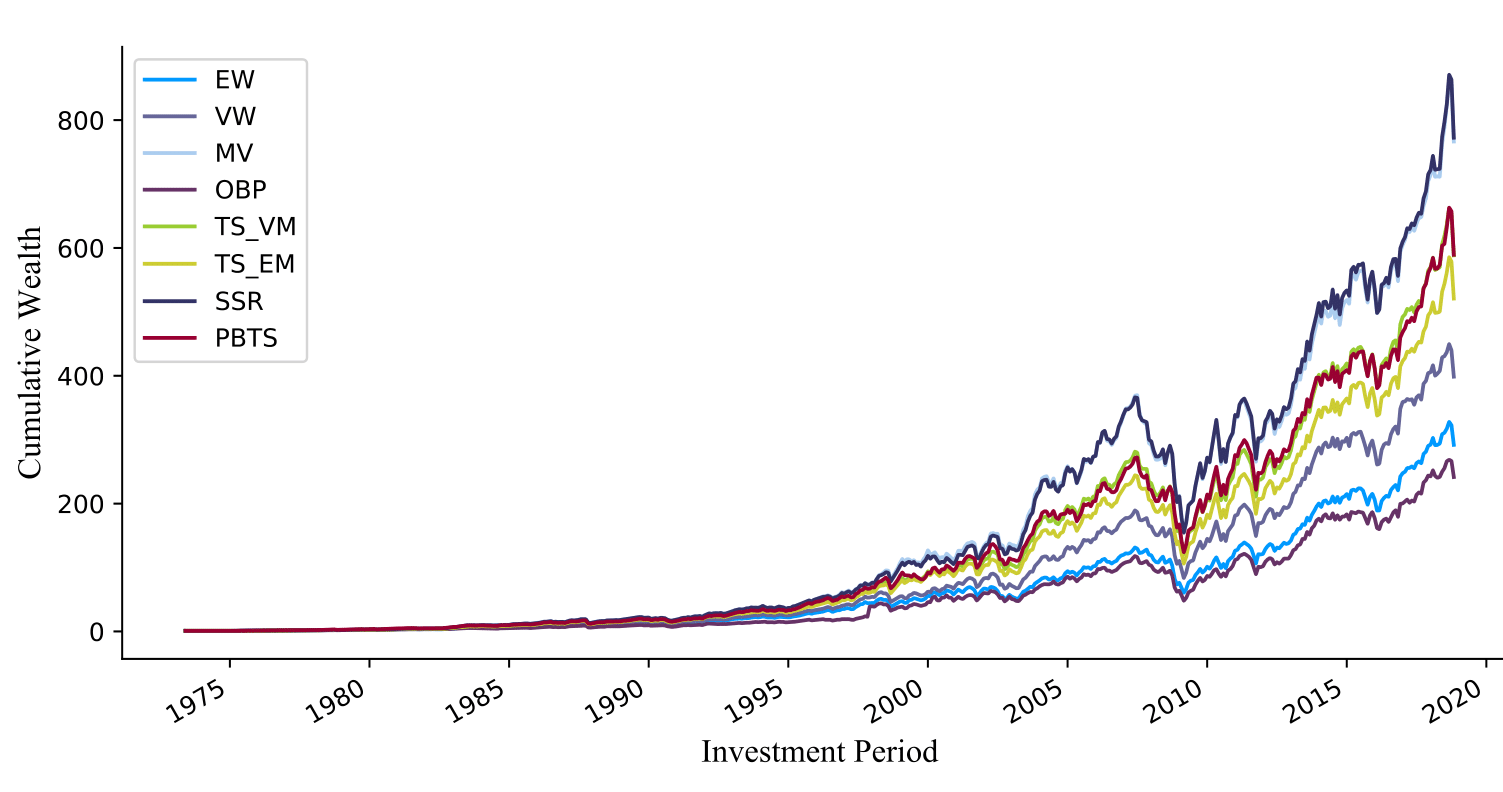}
\end{minipage}%
}%
\subfigure[FF49]{
\begin{minipage}[t]{0.33\linewidth}
\centering
\includegraphics[width=2.3in]{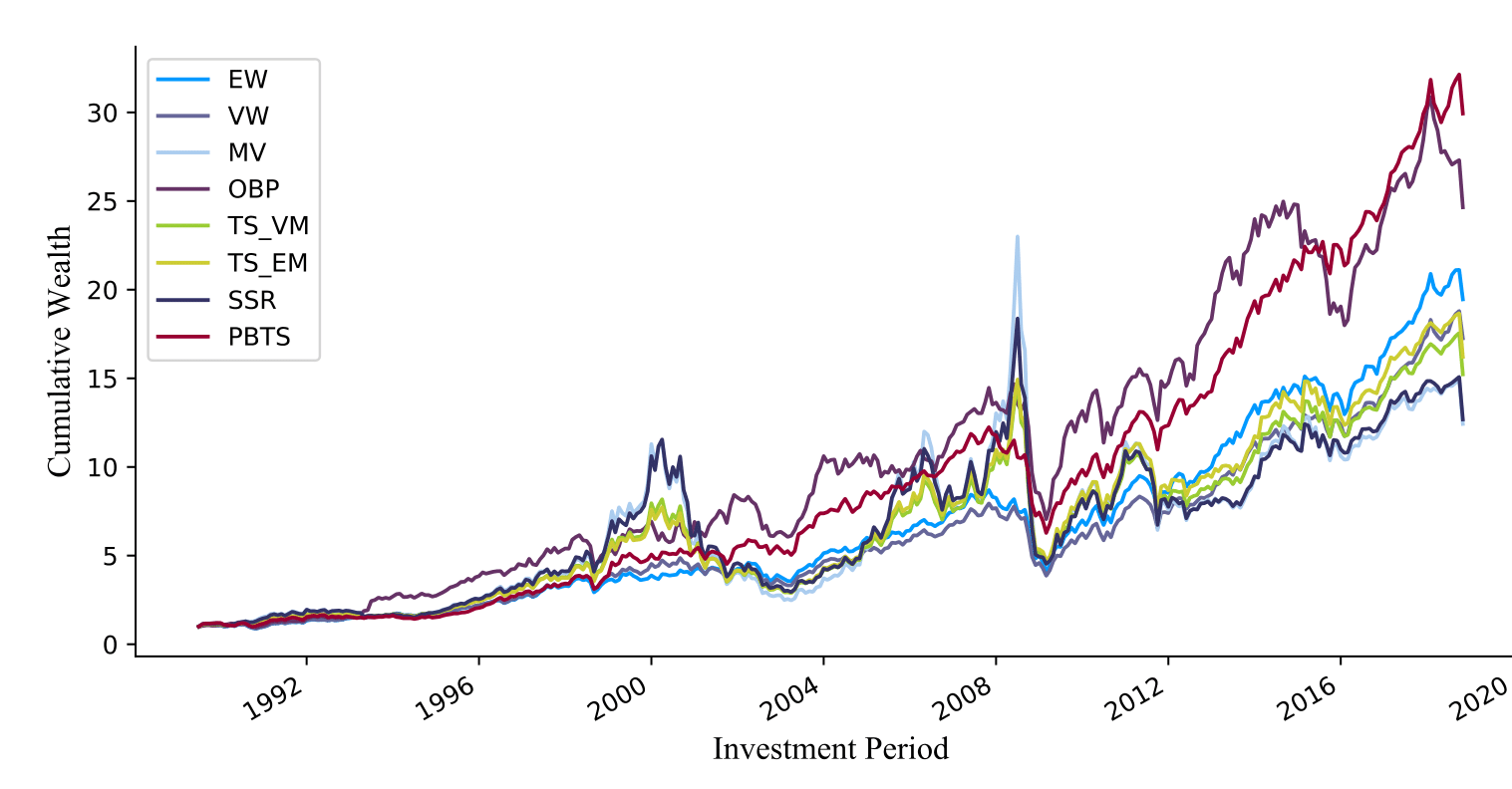}
\end{minipage}%
}%
\subfigure[FF100]{
\begin{minipage}[t]{0.33\linewidth}
\centering
\includegraphics[width=2.3in]{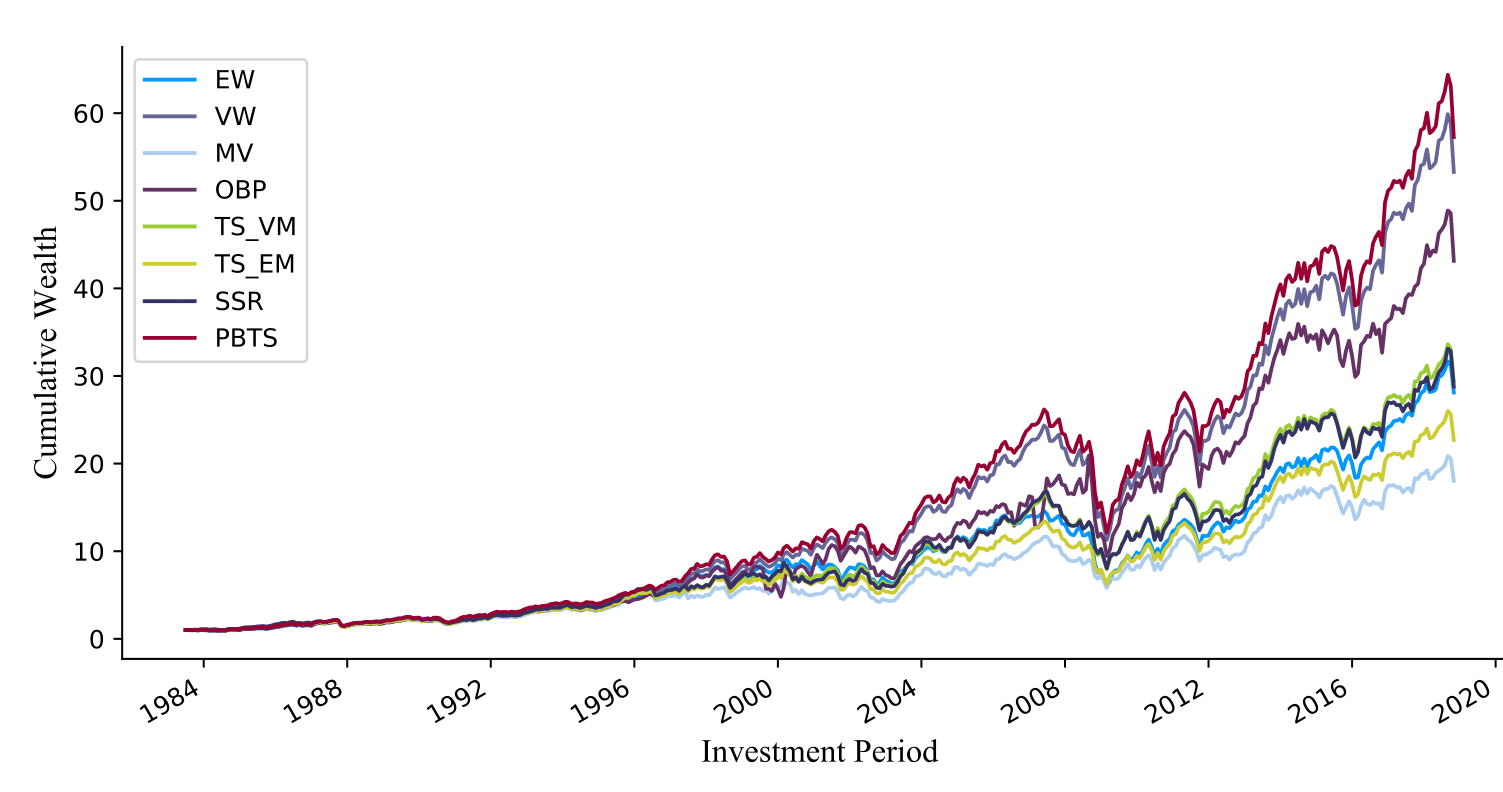}
\end{minipage}
}%

\subfigure[ETFs]{
\begin{minipage}[t]{0.33\linewidth}
\centering
\includegraphics[width=2.3in]{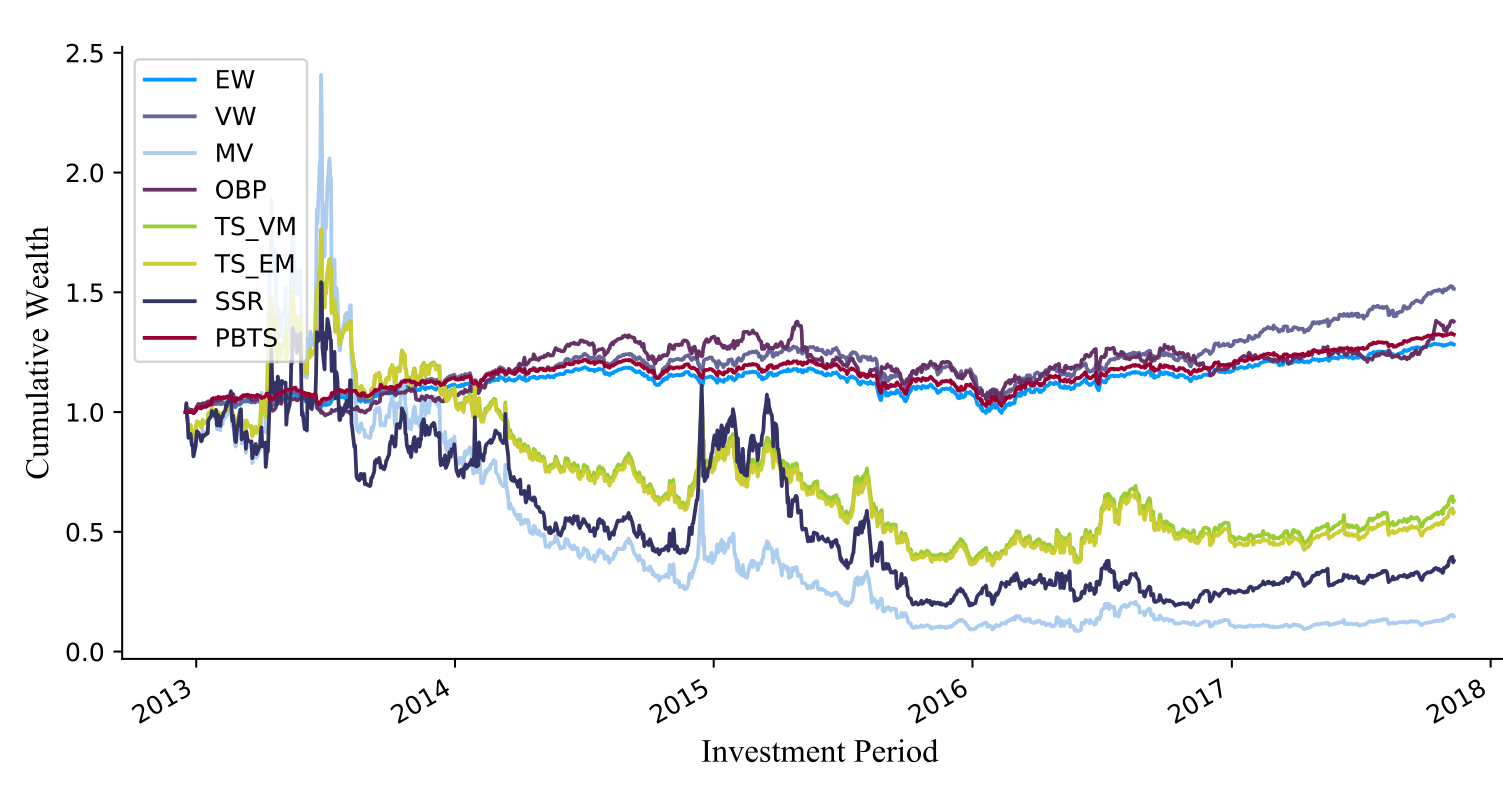}
\end{minipage}
}%
\subfigure[SP500]{
\begin{minipage}[t]{0.33\linewidth}
\centering
\includegraphics[width=2.3in]{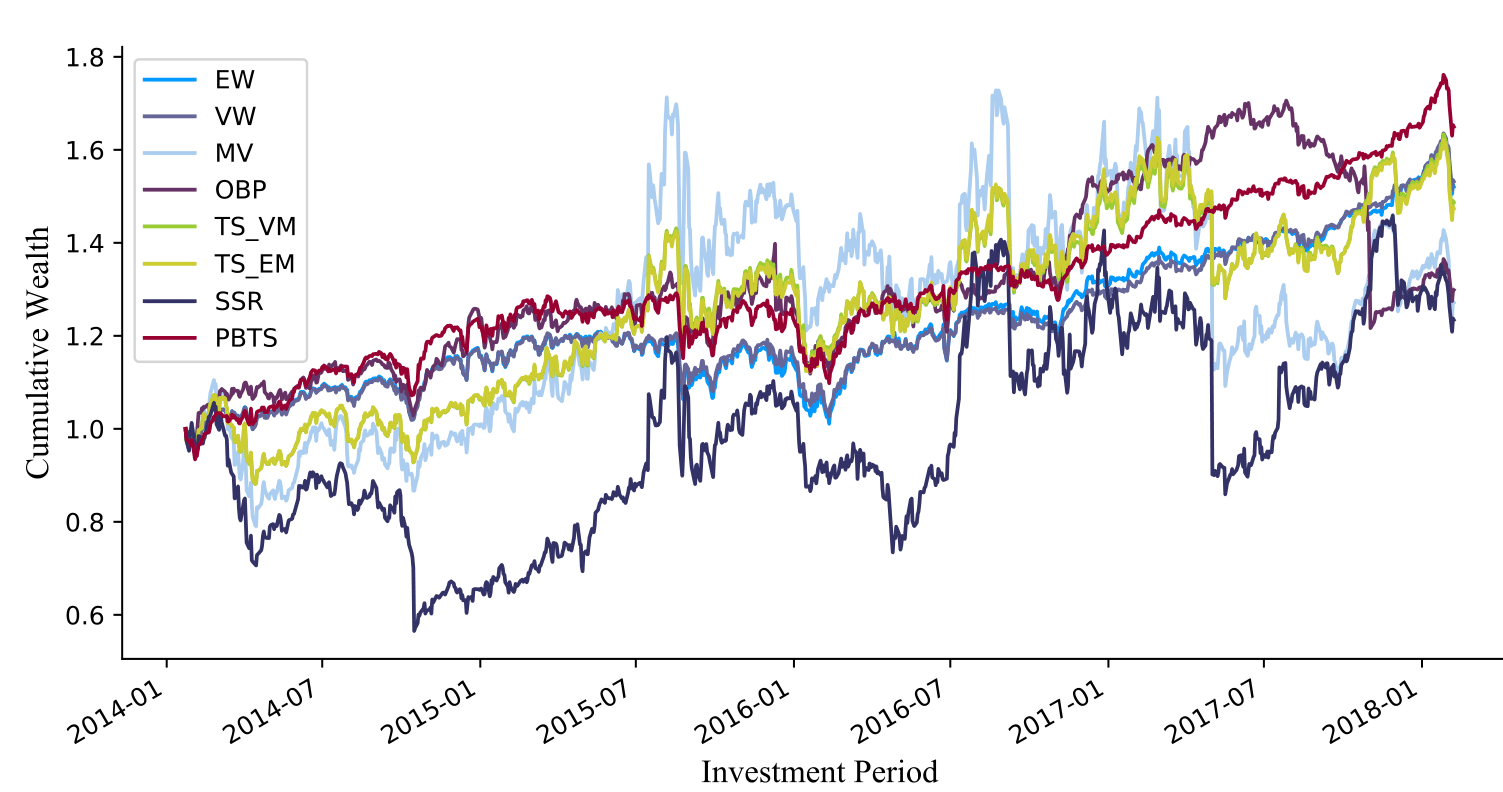}
\end{minipage}
}%
\centering
\caption{ The curves of cumulative wealth across the investment periods for different portfolios on (a) FF25, (b) FF49, (c) FF100, (d) ETFs, and (e) SP500 datasets.}
\label{fig:settings}
\end{figure*}

\subsection{Evaluation Metrics}

We use the standard criteria in finance \cite{brandt2010portfolio} to measure the performance of the portfolio strategy outside the training sample: (1) Sharpe Ratio; (2) Cumulative Wealth; (3) Maximum Drawdown; (4) Volatility.

\textbf{Sharpe Ratio (SR)} measures the return-to-risk ratio of a portfolio strategy and normalizes the return on the portfolio using its standard deviation. It is expressed as:
	\begin{equation}
	SR=\frac{\hat{\mu }}{\hat{\sigma }},\hat{\mu }=\frac{1}{m-\tau }\sum_{t=\tau+1}^{m}\mu _{t},\hat{\sigma }=\sqrt{\frac{1}{m-\tau }\sum_{t=\tau+1}^{m}(\mu _{t}-\hat{\mu })^2} \label{con:variance},
	\end{equation}
where $\mu _{t}=\mathbf{R_{t}}^{T}\mathbf{w_{t}}-1$.
	
SR is a comprehensive measure that combines both returns and risks into the evaluation, giving the return value of each risk of the portfolio.

\textbf{Cumulative Wealth (CW)} is a weighted cumulative return measuring the time at which each asset's revenue in a portfolio strategy begins to accumulate to the last calculated return. It is expressed as:\begin{equation}
CW= \prod_{t=\tau +1}^{m}\mathbf{R_{t}}^{T}\mathbf{w_{t}}.
\end{equation}

\textbf{Maximum Drawdown (MDD)} is the maximum amount of wealth reduction that a cumulative wealth has produced from its maximum value over time, expressed as:
\begin{equation}
MDD=\max _{t\in(\tau ,m) }(M_{t}-CW_{t}),M_t=\max _{k\in(\tau ,t) }CW_{k},
\end{equation}
where retracement $M_t - CW_t$ represents to the loss from the maximum wealth value $M_t$ during its operation to the time $t$, and $CW_t$ denotes to the cumulative wealth up to the time $t$. Since the sharp decline inevitably causes investors to panic and cause divestment, the maximum retracement is usually the primary risk measure for the money management industry.

\textbf{Volatility (VO)} is a quantitative risk metric for the investment industry. The calculation of portfolio volatility is related to the standard deviation in Equation (\ref{con:variance}). To measure the portfolio strategy with different weight adjustment frequencies, we calculate the annualized volatility using the following formula:
\begin{equation}
VO=\sqrt{H}\hat{\sigma},
\end{equation}
where $H$ is the number of times the weights are adjusted each year. In our experiment, $H = 12$ for the monthly datasets, and $H = 365$ for the daily datasets.

\subsection{Competing Portfolios}

To comprehensively assess the proposed method, we consider ten modern competing portfolios according to our literature review:

\textbf{Equally-weighted portfolio (EW):}
 EW is one of classic strategies, which. It has outperformed 14 sophisticated models across seven real-world datasets at monthly frequency of 2000 years \cite{demiguel2007optimal}. Therefore, EW is the first benchmark algorithm for portfolio research.

\textbf{Value-weighted portfolio (VW):}
VW is a strategy that imitates the market's passive portfolio, which is the same as the market index's volatility. It is also an important benchmark strategy.

\textbf{Mean-variance portfolio (MV):}
MV is one of our basic strategies based on Markowitz's theory and outperforms in different markets and time spans.

\textbf{Orthogonal Bandit portfolio (OBP):}
OBP constructs multiple assets by constructing orthogonal portfolios. It also uses the upper confidence bound bandit framework to derive the optimal portfolio strategy that represents the combination of passive and active investments  as per a risk-adjusted reward function \cite{shen2015Portfolio}.

\textbf{Portfolio Blending via Thompson Sampling (TS-EM, TS-VM):}
This strategy is applied by Thompson sampling to the portfolio field for mixing EW and MV as TS-EM, VW and MV as TS-VM \cite{shen2016portfolio}.

\textbf{Portfolio Selection via Subset Resampling (SSR):}
The SSR method estimates the parameters by re-sampling subsets of the original assets, and aggregates the subsets of the multiple constructs to obtain the portfolio of all assets \cite{shen2017portfolio}.

Generally, EW, VW, and MV are three portfolio strategy arms of PBTS, which should be compared with the hybrid model proposed in this paper. OBP, TS-EM, TS-VM, and SSR are the heuristic experiments of the model. They are well recognized as important portfolio strategies based on the exploration and exploitation problem. Therefore, to be more convincing, we also compare with these four models. 

\begin{table*}[ht]
\centering
\begin{tabular}{llllllllll}
\hline
Dataset  & Metrics & PBTS & EW & VW & MV & OBP & TS-EM & TS-VM & SSR \\
\hline
\multirow{4}*{FF25}   & SR  & \textbf{22.60} & 20.02 & 19.84 & 19.30 & 15.92  & 19.82 &19.93  & 19.08\\
~ & CW  & 589.41 & 291.93 & 398.67 & 766.58 & 241.60  & 588.41 & 520.81  & \textbf{772.46}\\
~ & MDD (\%)  & \textbf{43.83}  & 54.10 & 55.91 & 57.98 &59.41  & 57.07 & 56.60  & 58.49\\
~ & VO (\%)  & 17.71 & \textbf{17.51} & 17.68 & 18.20 & 22.03 & 17.71 & 17.60 & 18.41 \\
\cline{1-10}

\multirow{4}*{FF49}   & SR  & \textbf{24.20} & 23.15 & 23.22 & 11.77 &18.55  & 15.77 & 15.93 & 13.22\\
~ & CW  & \textbf{29.94} &19.46 & 17.26 & 12.43 & 24.65 & 15.23 & 16.21 & 12.68 \\
~ & MDD (\%)  & \textbf{38.30}  & 52.83 & 51.42 & 79.90 & 51.97  & 68.72 & 68.35  & 75.79\\
~ & VO (\%)  & \textbf{14.39}  & 15.10 & 15.05 & 29.76 &18.87  & 22.19 & 21.96  & 26.44\\
\cline{1-10}

\multirow{4}*{FF100}  & SR  & \textbf{21.76} & 20.71	& 21.43 & 19.21&15.81  & 20.85 & 20.62 & 20.21\\
~ & CW  & \textbf{57.27} & 28.12 & 53.28 & 18.04 &43.14  & 29.42 & 22.69  & 28.78\\
~ & MDD (\%)  & \textbf{30.76}  & 58.73 & 53.72 & 50.26 & 54.80  & 51.80 & 53.29 & 52.38 \\
~ & VO (\%)  & \textbf{16.06}  & 16.88 & 16.33 & 18.18 &22.16  & 16.77 & 16.94  & 17.30\\
\cline{1-10}

\multirow{4}*{ETFs}  & SR  & 194.49 & \textbf{197.22} & 147.67 & 17.93 & 70.94  & 31.38 & 31.42 & 19.40\\
~ & CW  & 1.15 & 1.28 & 1.51 & 0.15 & \textbf{1.88}  & 0.63 & 0.58 & 0.38\\
~ & MDD (\%)  & \textbf{15.40}  & 16.20 & 18.46 & 96.44 & 23.71  & 78.77 & 79.58  & 88.09\\
~ & VO (\%)  & 9.83  & \textbf{9.69} & 12.94 & 106.56 & 26.95  & 60.89 & 60.81 &  98.54\\
\cline{1-10}

\multirow{4}*{SP500}  & SR  & 126.88 & 124.49 & \textbf{127.56} & 41.27 & 52.54 & 66.71 & 66.56 & 39.77\\
~ & CW  & \textbf{1.65} & 1.52 & 1.53 & 1.27 &1.30  & 1.49 & 1.47 & 1.32\\
~ & MDD (\%)  & \textbf{14.97}  & 16.41 & 14.97 & 36.81 & 41.09 & 20.82 & 21.24  & 46.49\\
~ & VO (\%)  & \textbf{15.06}  & 15.35 & 14.98 & 46.32 & 36.38  & 28.65 & 28.72 & 48.07 \\
\hline
\end{tabular}
\caption{ Performance of portfolio strategies}
\label{tab:performance}
\end{table*}

\subsection{Parameter Settings}

We use the ``rolling range" setting proposed by DeMiguel \cite{demiguel2007optimal}. In regard to the model proposed in this paper, we set the sliding window as $\tau =120$. For the parameter $c$ of the PBTS, we utilize cross validation to establish the optimal parameters. And for the parameters of other comparison algorithms, we use the parameter settings recommended in the relevant studies.

\subsection{Results and Analysis}

\textbf{Results} Table \ref{tab:performance} summarize portfolio performance evaluated by the SR, CW, MDD, and VO for all the tested benchmarks, respectively. From the comparisons of the various methods, the values in bold represent the winners' performance. The proposed PBTS method achieves a better performance in most of the cases. On the one hand, for the SR, the results of the PBTS are in the first echelon on all datasets, with a slightly lower EW on the ETFs dataset as well as less than VW on the SP500 dataset. This indicates that the PBTS basically has a better return-to-risk ratio. For the absolute return indicator, we use Figure \ref{fig:settings}  to reflect the change in earnings over time. PBTS outperforms other methods on most datasets, only below the MV and SSR on the FF25 dataset and lower than VW and OBP on the ETFs dataset. However, the OBP and SSR method has large fluctuations on other datasets. As well, the robustness is lower than the PBTS method. On the other hand, PBTS performs better on the risk indicators. As summarized in Table \ref{tab:performance}, the MDD of PBTS is the smallest; while PBTS's VO is lower than EW in the ETFs dataset, which is usually the lowest VO in the classic strategies, and is superior to other comparison methods in other datasets. 

\textbf{Analysis} In summary, through the performance of the three long-term FF datasets, we believe that the PBTS method has an outstanding performance in terms of long-term performance. This is consistent with the goal of PBTS  of achieving excellent long-term returns. However, in the mid-term high frequency situation, through the ETF and SP500 datasets, we realize that PBTS is not robust enough and depends on the performance of basic strategies. 

\begin{figure}[t]
  \centering
  \includegraphics[width=1\linewidth]{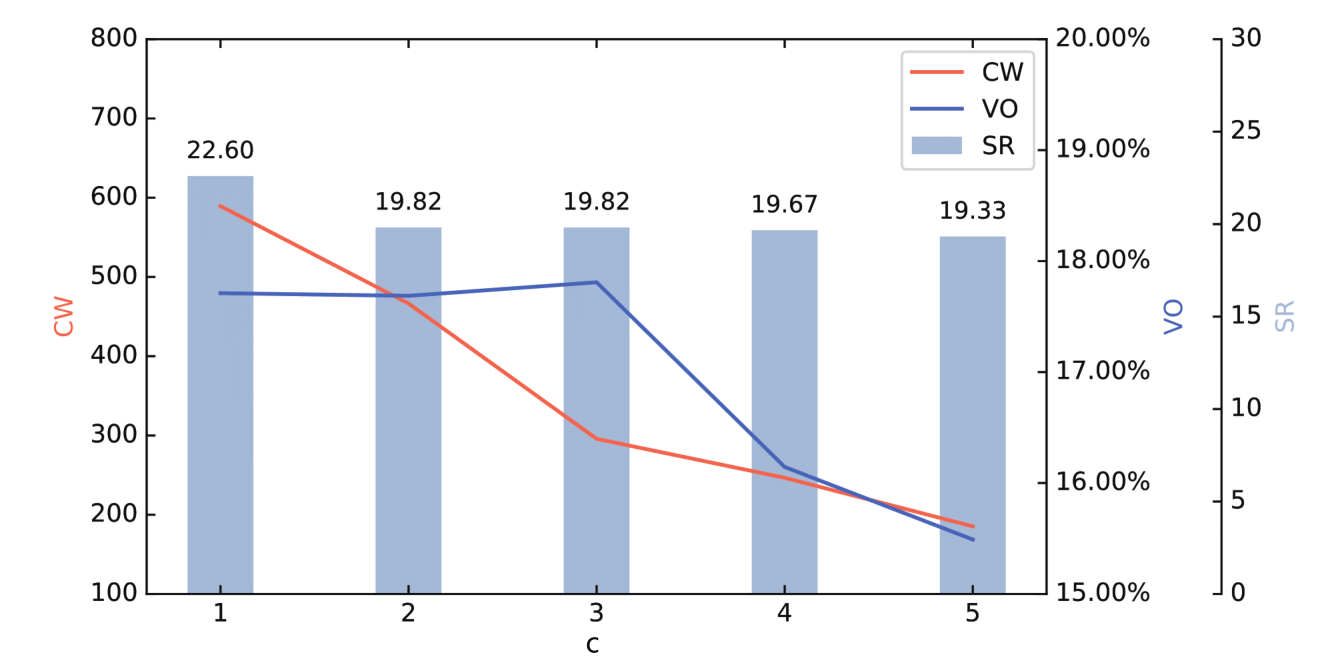}
  \caption{The effect of different $c$ on the performance of PBTS in FF25 dataset} 
  \label{fig:diffc} 
\end{figure}

\textbf{Parameter effect analysis} We analyze the performance of PBTS in the case of different $c$, as shown in Figure \ref{fig:diffc}. In FF25 dataset, as $c$ becomes larger, the volatility decreases, but the cumulative wealth decreases. This is consistent with our hypothesis that the larger the $c$, the lower the user's risk reference and the lower the risk that can be borne, but the return is reduced.

\section{Conclusions and Future Work}

In this paper, we constructed the portfolio selection problem into a multi-armed bandit problem, wherein we used the classic portfolio strategies as the strategic arms to form a dynamic portfolio strategy with multiple cycles to adapt for different periods. Moreover, we devise a reward function based on the user's investment risk preference to judge the standard and select the optimal arm of each period via Thompson sampling. Our algorithm could appropriately balance the benefits and risks well and achieve higher returns by controlling risk. 

In the future work, we will consider the correlation between the strategic arms and the impact of the previous selection path on the next choice. Also, the actual status of financial scenarios such as transaction fee, tax, and dividend should be considered as factors to build a portfolio strategy that is more consistent with the real scenario.



\bibliographystyle{named}
\bibliography{ijcai19}

\end{document}